\documentclass[10pt,twocolumn,letterpaper]{article}

\usepackage{iccv}
\usepackage{times}
\usepackage{epsfig}
\usepackage{graphicx}
\usepackage{amsmath}
\usepackage{amssymb}

\usepackage[pagebackref=true,breaklinks=true,letterpaper=true,colorlinks,bookmarks=false]{hyperref}

\iccvfinalcopy % *** Uncomment this line for the final submission

\def\iccvPaperID{6} % *** Enter the ICCV Paper ID here

\ificcvfinal\fi

\begin{document}

%%%%%%%%% TITLE
\title{Matrix Nets: A New Deep Architecture for Object Detection}

\author{Abdullah Rashwan\\
University of Waterloo\\
Vector Institute\\
{\tt\small arashwan@uwaterloo.ca}
\and
Agastya Kalra\\
Akasha Imaging Corp. \\
University of Waterloo\\
{\tt\small agastya@akasha.im} % \\ \tt\small a6kalra@uwaterloo.ca}
\and
Pascal Poupart\\
University of Waterloo\\
Vector Institute\\
{\tt\small ppoupart@uwaterloo.ca}}

\maketitle
\ificcvfinal\thispagestyle{empty}\fi
 \begin{figure}[t]
\begin{center}
\includegraphics[width=0.65\linewidth]{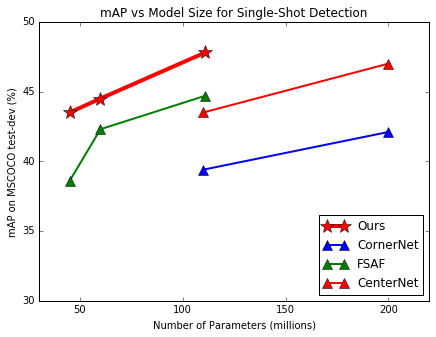}
\vspace{-0.5cm}
\end{center}
   \caption{Model size (millions of parameters) vs accuracy (average precision) reported on MSCOCO test-dev for single-shot detectors. Enabled by our MatrixNet backbone, our model outperforms all other single-shot architectures while being smaller in size.}
\label{fig:parameters_map}
\vspace{-0.5cm}
\end{figure}
%%%%%%%%% ABSTRACT
\begin{abstract}
We present Matrix Nets ($x$Nets), a new deep architecture for object detection. $x$Nets map objects with different sizes and aspect ratios into layers where the sizes and the aspect ratios of the objects within their layers are nearly uniform. Hence, $x$Nets provide a scale and aspect ratio aware architecture. We leverage $x$Nets to enhance key-points based object detection. Our architecture achieves mAP of 47.8 on MS COCO, which is higher than any other single-shot detector while using half the number of parameters and training 3x faster than the next best architecture.
\end{abstract}
\begin{figure}[t]
\begin{center}
\includegraphics[width=0.8\linewidth]{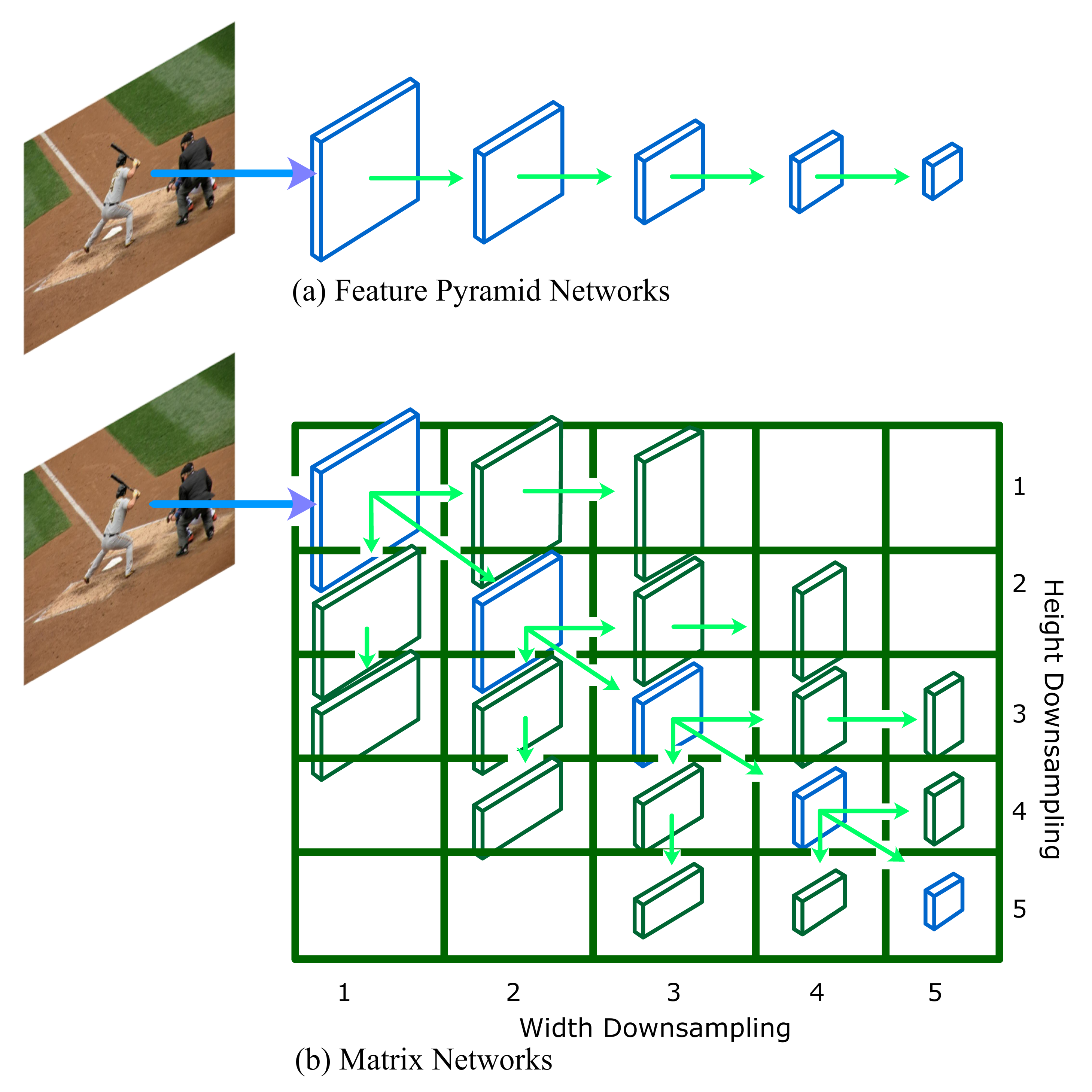}
\vspace{-0.5cm}
\end{center}
   \caption{(a) Shows the original FPN architecture \cite{lin2017feature}, where there are different output layers are assigned at each scale. Note we do not show the skip connections for the sake of simplicity. (b) Shows the MatrixNet architecture, where the 5 FPN layers are viewed as the diagonal layers in the matrix. We fill in the rest of the matrix by downsampling these layers.}
\label{fig:xnets}
\vspace{-0.5cm}
\end{figure}
%%%%%%%%% BODY TEXT
\section{Introduction}
Object detection is one of the most widely studied tasks in computer vision with many applications to other vision tasks such as object tracking, instance segmentation, and image captioning. Object detection architectures can be sorted into two categories; single-shot detectors \cite{lin2017focal, law2018cornernet} and two-stage detectors \cite{li2019scale}. Two-stage detectors leverage a region proposal network to find a fixed number of object candidates, then a second network is used to predict a score for each candidate and to refine its bounding box.

Single-shot detectors can also be split into two categories; anchor based detectors \cite{lin2017focal, zhu2019feature} and key-point based detectors \cite{law2018cornernet, duan2019centernet}. Anchor-based detectors contain many anchor boxes and then predict offsets and classes for each template. The most famous anchor-based architecture is RetinaNet \cite{lin2017focal}, which proposed the focal loss function to help correct for the class imbalance of positive to negative anchor boxes. The highest performing anchor-based detector is FSAF \cite{zhu2019feature}. FSAF ensembles the anchor-based output with an anchor free output head to further improve performance.

On the other hand, key-point based detectors predict top-left and bottom-right corner heat-maps and match them together using feature embeddings. The original key-point based detector is CornerNet \cite{law2018cornernet}, which leverages a special corner pooling layer to accurately detect objects of different sizes. Since then, CenterNet \cite{duan2019centernet}
substantially improved CornerNet architecture by predicting object centers along with corners.

Detecting objects at different scales is a major challenge for object detection. One of the biggest advancements in scale aware architectures is Feature Pyramid Networks (FPNs), introduced by Lin et al. \cite{lin2017feature}. FPNs were designed to be scale invariant by having multiple layers with different receptive fields so that objects are mapped to layers with relevant receptive fields. Small objects are mapped to earlier layers in the pyramid, and larger objects are mapped to later layers. Since the size of the objects relative to the downsampling of the layer are kept nearly uniform across pyramid layers, a single output sub-network can be shared across all layers. Although FPNs provided an elegant way for handling objects of different sizes, they didn't provide any solution for objects of different aspect ratios. A high tower, a giraffe, or a knife introduce a design difficulty for FPNs: Does one map these objects to layers according to their width or height? Assigning the object to a layer according to its larger dimension would result in loss of information along the smaller dimension due to aggressive downsampling, and vice versa. To solve this, we introduce Matrix Networks, a new scale and aspect ratio aware CNN architecture. $x$Nets, as shown in Fig.~\ref{fig:xnets}, have several matrix layers, each layer handles an object of specific size and aspect ratio. $x$Nets assign objects of different sizes and aspect ratios to layers such that object sizes within their assigned layers are close to uniform. This allows a square output convolution kernel to equally gather information about objects of all aspect ratios and scales. xNets can be applied to any backbone, similar to FPNs. We denote this by appending a "-X" to the backbone, i.e. ResNet50-X.

As an application, we use $x$Nets for key-point based object detection. While key-point based single-shot detectors are the current state-of-the-art \cite{duan2019centernet}, they have two limitations due to using a single output layer: 
they require very large, computationally expensive backbones, and special pooling layers for the model to converge. Second, they have difficulty accurately matching top-left, and bottom-right corners. To solve these limitations, we introduce keypoint-matrixnet (KP-$x$Net) architecture, an architecture that leverages $x$Net to achieve state-of-the-art results using ResNet-50, Resnet-101, and ResNeXt-101 backbones. We detect corners for objects of different sizes and aspect ratios using different matrix layers, and simplify the matching process by removing the embedding layer entirely and regressing the object centers directly. We show that KP-$x$Net outperforms all existing single-shot detectors by achieving 47.8\% mAP on the MS COCO benchmark.

The rest of the paper is structured as follows: Section 2 formalizes the idea of MatrixNets, while Section 3 discusses keypoint based object detection, and our method for applying MatrixNets for key-point based object detection. Section 4 covers experiments, results, and comparisons, and finally Section 5 is the conclusion.

\section{Matrix Nets}
\vspace{-0.25cm}
Matrix nets ($x$Nets) as shown in Fig. \ref{fig:xnets} model objects of different sizes and aspect ratio using a matrix of layers where each entry $i,j$ in the matrix represents a layer, $l_{i,j}$, with width down-sampling of $2^{i-1}$ and height down-sampling of $2^{j-1}$ with respect to the top left layer, $l_{1,1}$ in the matrix.  The diagonal layers are square layers of different sizes, equivalent to an FPN, while the off diagonal layers are rectangle layers, unique to $x$Nets. Layer $l_{1,1}$ is the largest layer in size, every step to the right cuts the width of the layer by half, while every step down cuts the height by half. For example $Width(l_{3,4}) =  0.5 Width(l_{3,3})$. Diagonal layers model objects with square-like aspect ratios, while off diagonal layers model objects with more extreme aspect ratios. Layers close to the top right or bottom left corners of the matrix model objects with very high or very low aspect ratios. Such objects are very rare, so these layers can be pruned for efficiency.

\begin{figure}[t]
\begin{center}
\includegraphics[width=1.0\linewidth]{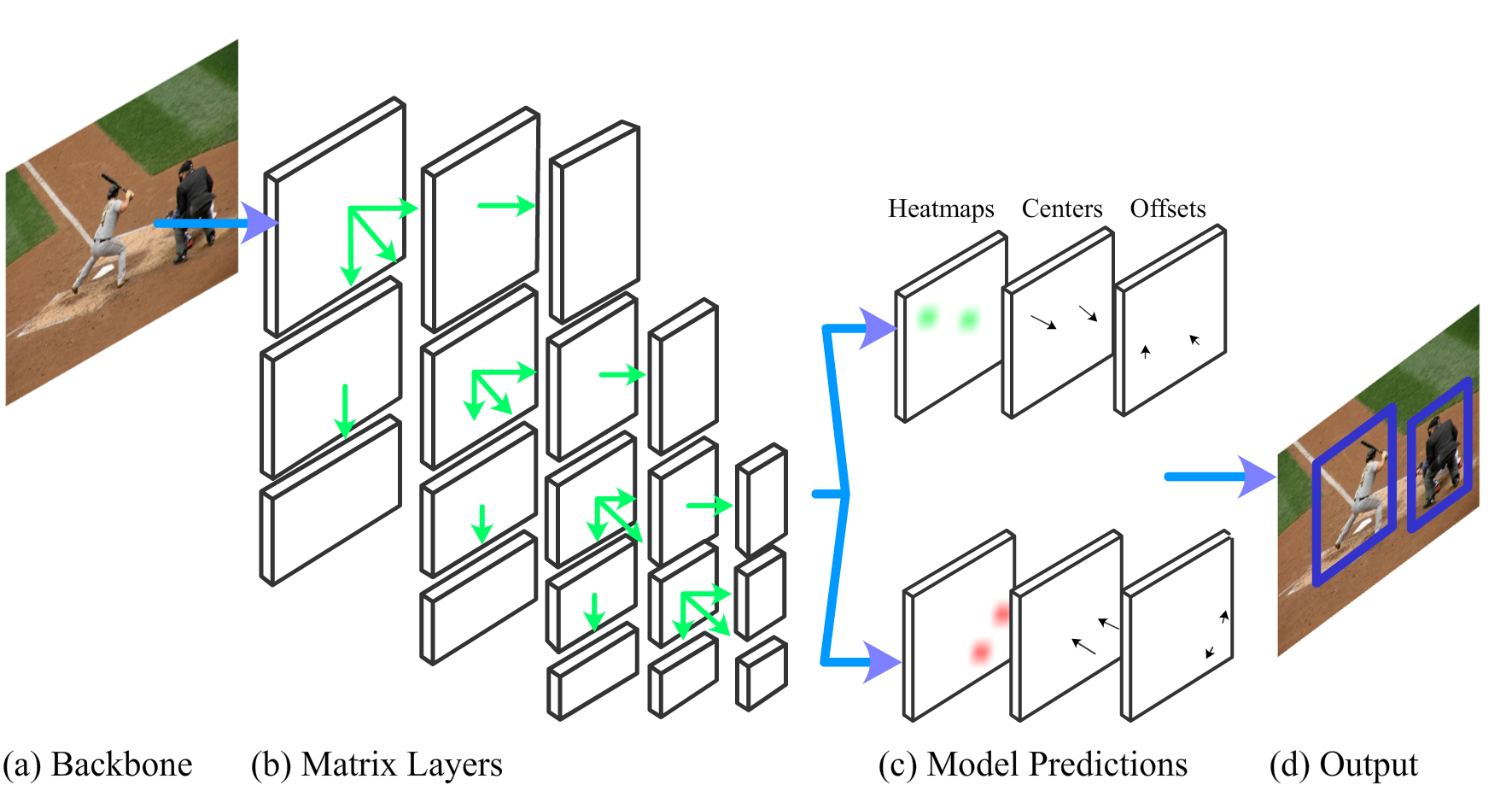}
\end{center}
\vspace{-0.25cm}
  \caption{The KP-$x$Net architecture.}
\label{fig:long}
\label{fig:KPxNet}
\vspace{-0.5cm}
\end{figure}
\subsection{Layer Generation}
\vspace{-0.25cm}
Generating matrix layers is a crucial step since it impacts the number of model parameters. The more parameters, the more expressive the model and the harder the optimization problem is, hence we chose to introduce as few new parameters as possible. The diagonal layers can be obtained from different stages of the backbone or using a feature pyramid backbone  \cite{lin2017feature}. The upper triangular layers are obtained by applying a series of shared 3x3 convolutions with stride 1x2 on the diagonal layers. Similarly, the bottom left layers are obtained using shared 3x3 convolutions with stride 2x1. The parameters are shared across all down-sampling convolutions to minimize the number of new parameters.

\subsection{Layer Ranges}
\vspace{-0.25cm}
Each layer in the matrix models objects of certain widths and heights, hence we need to define the range of widths and heights of objects assigned to each layer in the matrix. The ranges need to reflect the receptive field of the feature vectors of the matrix layers. Each step to the right in the matrix effectively doubles the receptive field in the horizontal dimension, and each step down doubles the receptive field in the vertical dimension. Hence, the range of the widths or heights needs to be doubled as we advance to the right or down in the matrix. Once the range for the first layer $l_{1,1}$ is defined, we can generate the ranges for the rest of the matrix layers using the above rule. For example, if the range for layer $l_{1,1}$ is $H\in [24px,48px]$, $W\in [24px,48px]$, the range for layer $l_{1,2}$ will be $H\in [24,48]$, $W\in [48,96]$. 

Objects on the boundaries of these ranges could destabilize training since layer assignment would change if there's a slight change in object size. To avoid this problem, we relax the layer boundaries by extending them in both directions. This is accomplished by multiplying the lower end of the range by a number less than one, and the higher end by a number greater than one, in all our experiments, we use 0.8, and 1.3 respectively.

\subsection{Advantages of Matrix Nets}
The key advantage of Matrix Nets is they allow a square convolutional kernel to accurately gather information about different aspect ratios. In traditional object detection models, such as RetinaNet, a square convolutional kernel is required to output boxes of different aspect ratios and scales. This is counter-intuitive since boxes of different aspect ratios and scales require different contexts. In Matrix Nets, the same square convolutional kernel can be used for detecting boxes of different scales and aspect ratios since the context changes in each matrix layer. Since object sizes are nearly uniform within their assigned layers, the dynamic range of the widths and heights is smaller compared to other architecture such as FPNs. Hence, regressing the heights and widths of objects becomes an easier optimization problem. Finally MatrixNets can be used as a backbone to any object detection architecture, anchor-based or keypoint-based, one-stage or two-stage detectors.

\section{Key-point Based Object Detection}
CornerNet \cite{law2018cornernet} was proposed as an alternative to anchor-based detectors, CornerNet predicts a bounding box as a pair of corners: top-left, and bottom-right. For each corner, CornerNet predicts heatmaps, offsets, and embeddings. Top-left, and bottom-right corner candidates are extracted from the heatmaps. Embeddings are used to group the top-left, and bottom-right corners that belong to the same object. Finally, offsets are used to refine the bounding boxes producing tighter bounding boxes. This approach has three main limitations. 

(1) CornerNet handles objects from different sizes and aspect ratios using a single output layer. As a result, predicting corners for large objects presents a challenge since the available information about the object at the corner location isn't always available with regular convolutions. To solve this challenge, CornerNet introduced the corner pooling layer that uses a max operation on the horizontal and vertical dimensions. The top left corner pooling layer scans the entire right bottom image to detect any presence of a corner. Although, experimentally they show that corner pooling stabilizes the model, we know that max operations lose information. For example, if two objects share the same location for the top edge, only the object with the max features will contribute to the gradient. So, we can expect to see false positive predictions due to corner pooling layers.

(2) Matching the top left and bottom right corners is done with feature embeddings. There are two problems that arise from using embeddings in this setting. First, the pairwise distances need to be optimized during the training, so as the number of objects in an image increases, the number of pairs increases quadratically, which affects the scalability of the training when dealing with dense object detection. The second problem is learning embeddings themselves. CornerNet tries to learn the embedding for each object corner conditioned on the appearance of the other corner of the object. Now, if the object is too big, the appearance of both corners can be very different due to the distance between them, as a result the embeddings at each corner can be different as well. Also, if there are multiple objects in the image with similar appearance, the embeddings for their corners will likely be similar. This is why we saw examples where CornerNet merged persons, or traffic lights together.

(3) As a result of the previous two problems, CornerNet is forced to use the Hourglass-104 backbone to achieve state-of-the-art performance. This has over 200M parameters, very slow and unstable training, requiring 10 GPUs with 12GB memory to ensure a large enough batch size for stable convergence.

\subsection{Key-point Based Object Detection Using Matrix Nets}
\vspace{-0.25cm}
Fig.~\ref{fig:KPxNet} shows our proposed architecture for keypoint based object detection, KP-$x$Net. KP-$x$Net consists of 4 stages. (a-b) We use a $x$Net backbone as defined in Section 2. (c) Using a shared output sub-network, for each matrix layer we predict the top-left and bottom-right corner heatmaps, corner offsets, and center predictions for objects within their layers. (d) We match corners within the same layer using the center predictions, and then combine the outputs of all layers with soft non-maximum suppression to achieve the final output.

\textbf{Corner Heatmaps}
Using $x$Nets ensures that the context required for objects within a layer is bounded by the receptive field of a single feature map in that layer. As a result, corner pooling is no longer needed, regular convolutional layers can be used to predict the heatmaps for the top left and bottom right corners. Similar to CornerNet, we use focal loss to deal with unbalanced classes.

\textbf{Corner Regression}
Due to image downsampling, refining the corners is important to have tighter bounding boxes. When scaling down a corner to $x$, $y$ location in a layer, we predict the offsets so that we can scale up the corner to the original image size without losing precision. We keep the offset values between $-0.5$, and $0.5$, and we use smooth L1 loss to optimize the parameters.

\textbf{Center Regression}
Since the matching is done within each individual matrix layer,  the width and height of the object is guaranteed to be within a certain range. The center of the object can be regressed easily because the range for the center is small. In CornerNet, the dynamic range for the centers is large, trying to regress centers in a single output layer would probably fail. Once the centers are obtained, the corners can be matched together by comparing the regressed center to the actual center between the two corners. During the training, center regression scales linearly with the number of objects in the image compared to quadratic growth in the case of learning embeddings. To optimize the parameters, we use smooth L1 loss.

KP-$x$Net solves problem (1) of CornerNets because all the matrix layers represent different scales and aspect ratios rather than having them all in a single layer. This also allows us to get rid of the corner pooling operation. (2) is solved since we no longer predict embeddings, instead we regress centers directly. By solving the first two problems of CornerNets, we will show in the experiments that we can achieve higher results than CornerNet using a smaller network, and less of computational resources.
\vspace{-0.25cm}
\section{Experiments}
\vspace{-0.25cm}
We train all of our networks on a server with Titan XP GPUs. We use a batch size of 20, that requires 3 GPUs for resnet50-X, and 4 GPUs for resnet101-X, and ResNeXt101-X. For our final ResNeXt101-X experiment we train on an AWS p3.16xlarge instance with 8 V100 GPUs to allow for a larger batch size of 55. This improves performance by ~0.7\% mAP. During the training, we use crops of sizes 512x512, and we use standard scale jitter of 0.6-1.5 and a custom cutout \cite{devries2017improved} implementation. For optimization, we use the Adam optimizer and set an initial learning rate of 5e-5, and cut it by 1/10 after 60 epochs, training for a total of 80 epochs. For our matrix layer ranges, we set $l_{1,1}$ to be [24px-48px]x[24px-48px] and then scale the rest as described in Section 2.2. At test time, we resize the image so that the max side of the image is 900. We trained our model on MS COCO 'trainval35k' set (i.e., 80K training images and 35K validation images), and tested on the 'test-dev2017' set.  Using this setup, we achieve 41.7, 42.7, 44.7 mAP single scale, and 43.9, 44.8, and 47.8 mAP multi-scale.
\subsection{Comparisons}
\vspace{-0.25cm}
\begin{table}[]
\caption{Empirical results on MSCOCO for our method as compared to best results reported in other works.}
\begin{tabular}{@{}lll@{}}
 \hline
\textbf{Architecture}              & \textbf{Backbone}      & \textbf{mAP}  \\
 \hline
CornerNet \cite{law2018cornernet}                         & Hourglass-104           & 40.8          \\
CornerNet (Multi-Scale)   \cite{law2018cornernet}           & Hourglass-104           & 42.1          \\
RetinaNet  \cite{lin2017focal}                        & ResNeXt-101-FPN        & 40.8          \\
FSAF \cite{zhu2019feature}                              & ResNeXt-101-FPN        & 42.3          \\
FSAF (Multi-Scale) \cite{zhu2019feature}                & ResNeXt 101-FPN        & 44.6          \\
%CenterNet-1 \cite{zhou2019objects}                       & Hourglass-104          & 40.3          \\
%CenterNet-1 (Multi-Scale)  \cite{zhou2019objects}        & Hourglass 104          & 45.1          \\
CenterNet \cite{duan2019centernet}                       & Hourglass-104           & 44.9          \\
CenterNet (Multi-Scale) \cite{duan2019centernet}         & Hourglass-104           & 47.0          \\  \hline
KP-$x$Net                            & ResNeXt-101-X          & 44.7 \\ 
\textbf{KP-$x$Net (Multi-Scale)}     & \textbf{ResNeXt-101-X}   & \textbf{47.8}   \\ 
\hline
\end{tabular}

\label{table:results}
\vspace{-0.5cm}
\end{table}

As shown in Table~\ref{table:results}, our final model, KP-$x$Net (Multi-Scale) with ResNext-101-X backbone achieve higher mAP than the next best model and more than 5.7\% mAP over the original CornerNet architecture. The second best architecture, CenterNet (Multi-Scale), is trained with a backbone twice the size and 3x the training iterations, and 2x GPU memory. CenterNet ran 480k training iterations on the Hourglass-104 backbone whereas our model converged in 180k training iterations on the smaller ResNeXt-101-X backbone. Even RetinaNet takes 150 epochs to converge, which is 1.8x more than ours which takes only 80 epochs. 
  
We also compare our model with other models on different backbones based on the number of parameters. In Fig.~\ref{fig:parameters_map}, we show that KP-$x$Net outperforms all other architectures at all parameter levels. Results are obtained from each individual paper. We believe this is because KP-$x$Net uses a scale and aspect ratio aware architecture.  
\vspace{-0.25cm}
\section{Conclusion}
\vspace{-0.15cm}
In this work, we introduced MatrixNet, a scale and aspect ratio aware architecture for object detection. We showed how to use MatrixNets to solve fundamental limitations of keypoints object detection. Our model achieves the state-of-the-art accuracy on MS COCO among single shot detectors.
\vspace{-0.25cm}
{\small
\bibliographystyle{ieee}
\bibliography{main}
}
\end{document}